\documentclass{article}
\usepackage{amssymb}
\usepackage{amsmath}
\usepackage{array}
\usepackage{algorithmic}
\usepackage{graphicx}
\usepackage{booktabs}
\usepackage{multirow}
\usepackage{pifont}
\usepackage{makecell}
\usepackage[table,xcdraw]{xcolor} 
\usepackage{algorithm}
\usepackage{xcolor}

\usepackage[preprint]{corl_2025} 
\usepackage{booktabs} 
\usepackage{float}
\usepackage{makecell}
\usepackage{amsmath} 

\usepackage{graphicx}
\usepackage{multirow}
\usepackage{wrapfig} 
\usepackage{subcaption} 

\title{EndoVLA: Dual-Phase Vision-Language-Action Model for Autonomous Tracking in Endoscopy} 

\author{
  \textbf{Chi Kit Ng}$^{1}$\thanks{Equal contribution.} \quad
  \textbf{Long Bai}$^{1}$\footnotemark[1] \quad
  \textbf{Guankun Wang}$^{1}$\footnotemark[1] \quad
  \textbf{Yupeng Wang}$^{1}$ \quad
  \textbf{Huxin Gao}$^{1}$ \\
  \textbf{Kun Yuan}$^{1,2}$ \quad
  \textbf{Chenhan Jin}$^{1}$ \quad
  \textbf{Tieyong Zeng}$^{1}$ \quad
  \textbf{Hongliang Ren}$^{1}$\thanks{Corresponding author: hlren@ee.cuhk.edu.hk} \\
  $^1$ The Chinese University of Hong Kong \quad
  $^2$ Technical University of Munich
}

\begin{document}
\maketitle
\begin{abstract}
In endoscopic procedures, autonomous tracking of abnormal regions and following circumferential cutting markers can significantly reduce the cognitive burden on endoscopists. However, conventional model-based pipelines are fragile — each component (e.g., detection, motion planning) requires manual tuning and struggles to incorporate high-level endoscopic intent, leading to poor generalization across diverse scenes. Vision-Language-Action (VLA) models, which integrate visual perception, language grounding, and motion planning within an end-to-end framework, offer a promising alternative by semantically adapting to surgeon prompts without manual recalibration. Despite their potential, applying VLA models to robotic endoscopy presents unique challenges due to the complex and dynamic anatomical environments of the gastrointestinal (GI) tract. To address this, we introduce EndoVLA, designed specifically for continuum robots in GI interventions. Given endoscopic images and surgeon-issued tracking prompts, EndoVLA performs three core tasks: (1) polyp tracking, (2) delineation and following of abnormal mucosal regions, and (3) adherence to circular markers during circumferential cutting. To tackle data scarcity and domain shifts, we propose a dual-phase strategy comprising supervised fine-tuning on our EndoVLA-Motion dataset and reinforcement fine-tuning with task-aware rewards. Our approach significantly improves tracking performance in endoscopy and enables zero-shot generalization in diverse scenes and complex sequential tasks.
\end{abstract}

\section{Introduction}
Endoscopic procedures is the gold standard for robotic-assisted minimally invasive surgery that rely on flexible robotic endoscopes to navigate the long and complex GI tract and to locate and treat lesions~\cite{gaoijrr,maple2015endoscopic,lee2021easyendo}. A critical capability during these interventions is tracking of endoluminal targets—such as polyps, lesion boundaries, or circumferential cutting markers—to guide precise manipulation. However, due to the non-intuitive nature of endoscope manipulation, surgeons must undergo extensive and systematic training to develop the required skills~\cite{zhang2020learning}. The success of endoscopic surgery highly depends on the surgeon's clinical experience, which can be influenced by limitations of preoperative and intraoperative imaging, restricted visual fields, and subjective biases, etc. These challenges can impair the accuracy of endoscopy planning and execution. Additionally, physical constraints such as hand tremors and fatigue may exacerbate procedural difficulty~\cite{lin2023end, hwang2020applying}.

\begin{figure}[t]
  \centering
  \includegraphics[width=\linewidth]{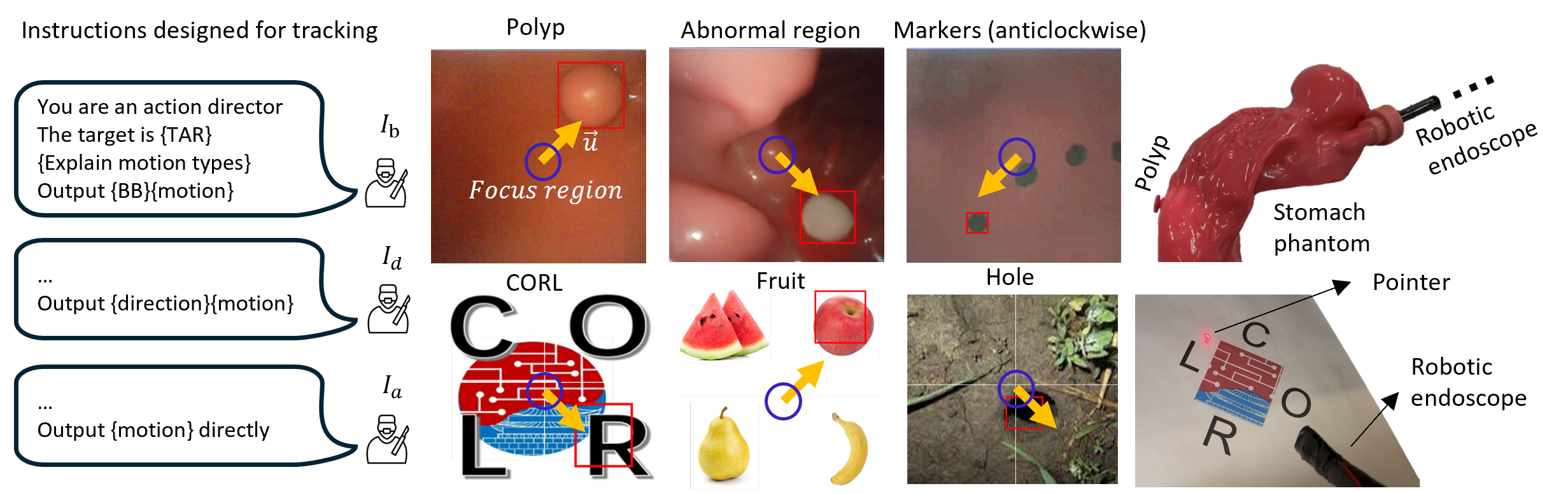}
  \caption{EndoVLA enables robust autonomous tracking in endoscopic procedures, demonstrating effective zero-shot generalization capabilities across general scenes and sequential tracking tasks.}
  \vspace{-2em} 
  \label{fig:EndoVLA-fig}
\end{figure}

Autonomous tracking capabilities for following abnormal regions and circumferential cutting markers could significantly reduce the cognitive burden on endoscopists~\cite{islam2019learning}. However, the GI environment is highly dynamic and unstructured, characterized by continuous tissue deformation and specular reflections that cause rapidly changing appearances. These factors render conventional vision-based tracking pipelines brittle, as they typically decompose the task into distinct perception, motion planning, and control modules—each requiring manual tuning and lacking the ability to generalize across diverse anatomical scenes and incorporate high-level surgical intent~\cite{leonard2018evaluation,fu2021future}. To tackle these constraints, learning-based strategies such as reinforcement learning have been proposed to address specific endoscopic tasks, aiming to improve precision and outcomes~\cite{long2023human,xu2021surrol,huang2023guided}. However, these approaches often require intensive data, detailed reward engineering, and extensive parameter tuning. They also lack generalizability across different endoscopic scenarios, making it difficult to address domain gaps in safety-critical in vivo tasks, thereby limiting their autonomy~\cite{li2024robonurse,moghani2024sufia}. 

Multimodal Large Language Models (MLLMs) have gained significant attention for the ability to effectively process textual instructions and visual inputs~\cite{achiam2023gpt,liu2023visual,wang2025endochat}. These models have demonstrated substantial potential in vision-language-action (VLA) paradigms for robotic planning, control, and execution~\cite{li2024robonurse,zitkovich2023rt,liu2024robomamba,arai2025covla,kim2024openvla}. MLLMs can directly interpret the surgeon's instructions through natural language (e.g., tracking the polyp), dynamically adjust task objectives as needed, and require minimal training or tuning. 
Despite these advantages, directly applying VLA frameworks to robotic endoscopy remains challenging due to the anatomical context varies significantly across patients and regions of the GI tract~\cite{moghani2024sufia,rudiman2021minimally}. While MLLMs are trained on general-purpose datasets, they still exhibit the semantic gap when applied to the complex and dynamic in vivo scenarios of endoscopic environments. Moreover, their open-ended nature can conflict with the deterministic and reliable behavior required in endoscopy~\cite{wang2025endochat}. Consequently, a customized solution is essential to ensure safe, accurate performance in this safety-critical in vivo environment~\cite{fan2024learn}.

To address these challenges, we propose EndoVLA, a dual-phase VLA framework designed specifically for autonomous tracking with continuum robotic endoscopes. Given endoscopic images and surgeon-issued tracking prompts, EndoVLA performs three core tracking tasks essential for autonomous assistance: polyp tracking, delineation and following of abnormal mucosal regions, and adherence to predefined circular markers during circumferential cutting. To achieve robust performance and generalization, our approach incorporates a novel dual-phase fine-tuning (DFT) strategy that combines supervised fine-tuning (SFT) with reinforcement fine-tuning (RFT) using task-aware verifiable rewards. Specifically, our contributions are as follows: (1) We introduce EndoVLA, an end-to-end VLA model tailored for the continuum endoscopic robot. We further employ the DFT approach, combining SFT and RFT for robust performance and generalization. (2) We construct the EndoVLA-Motion dataset for vision-language-kinematics modeling of endoscopic robots. (3) Extensive real-world experiments demonstrate that EndoVLA achieves state-of-the-art performance across three endoscopic tracking tasks while exhibiting remarkable zero-shot generalization to tracking tasks in general scenes and more challenging sequential tasks.
\section{Related Work}

\textbf{Multimodal Large Language Models in Robotics.}
Recently, there has been substantial advancement in the development of multimodal large language models tailored for robotic applications~\cite{brohan2022rt,o2024open,hu2023toward}. These approaches are trained on large-scale datasets, taking as input sequences of images from robots’ cameras with task instructions in natural language, and predicting motion outputs~\cite{schmidgall2024general}. VoxPoser~\cite{shridhar2023perceiver} introduces an LLM-guided perception system capable of grounding daily commands into manipulation primitives using visual information. Robotics Transformer 2~\cite{brohan2023rt} extended this concept with large-scale web and robotic data to support zero-shot generalization. Rather than relying on token-based action representations, Black et al.~\cite{black2024pi_0} adopt continuous action generation using flow matching, allowing precise, high-frequency control suitable for dexterous tasks. Although great progress has been achieved in robotics, these techniques often exhibit poor generalization to endoscopic domains, as MLLMs optimized for specific tasks tend to struggle when applied to more complex tasks in diverse endoscopic scenarios.

\textbf{Robot Learning for Automatic Surgical Robotics.}
Endoscopy-assisted surgical robotics has increasingly adopted reinforcement learning (RL) and imitation learning (IL) to achieve the precision and adaptability necessary for delicate clinical procedures. Deep RL has been employed to acquire complex surgical skills such as suturing, tissue manipulation, and needle handling, achieving notable improvements in accuracy and autonomy across simulated and real-world environments~\cite{qian2025deep}. To address the high data demands of RL, demonstration-guided RL integrates expert trajectories to accelerate convergence and reduce exploration in tasks like knot tying and percutaneous needle insertion~\cite{huang2023guided}. In parallel, sequence-based IL from surgical video data enables the learning of instrument motion patterns and semantic understanding directly from intraoperative recordings, advancing applications such as laparoscope guidance and endoscopic navigation under limited supervision~\cite{liu2024surgical}. Moreover, large-scale video-based IL models trained on da Vinci robotic surgery datasets have demonstrated near-surgeon-level proficiency in foundational tasks, including needle manipulation and tissue retraction~\cite{kim2024surgical}. While prior techniques have shown promising performance, their application to continuum endoscopy remains unexplored, and challenges like fine-grained control and real-time perception for continuum endoscopic robots have yet to be addressed.

\section{Robotic System and Dataset Collection}
\subsection{Endoscopic Robot System}

\begin{figure}[t]
  \centering
  \includegraphics[width=\linewidth]{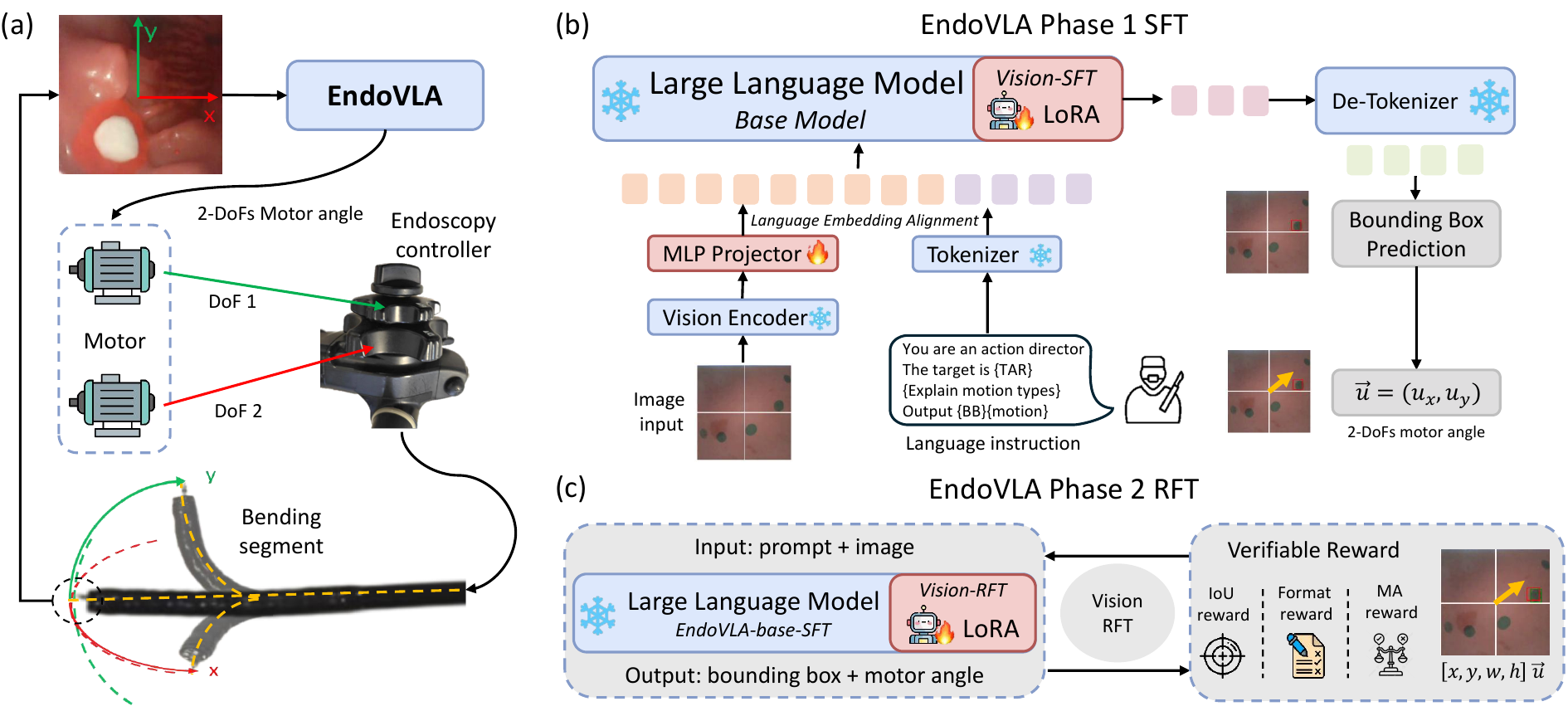}
  \caption{Overview of the setup of robotic endoscope and the DFT architecture of EndoVLA}
  \vspace{-1.5em} 
  \label{fig:robot_setup}
\end{figure}

To implement our proposed EndoVLA on an actual robotic system, we developed an endoscopic robot system for automatic control based on the Olympus endoscope. The commercial Olympus endoscope features two control knobs at its operating end, which regulate the 2-degree-of-freedom (2-DOFs) movement of the endoscopic end-effector. This 2-DOFs movement, as depicted in Figure \ref{fig:robot_setup} (a), allows the front segment of the robot to bend towards four perpendicular directions. By utilizing two motors to independently drive these knobs, the system achieves full coverage of movement within the imaging field, enabling real-time tracking of the target. The endoscopic system also provides a real-time video feed at 30 FPS. All experiments are conducted at a resolution of \(400 \times 400\) pixels, corresponding to the maximum output resolution of the endoscopic camera. Based on the output from EndoVLA, motion is converted to angle under the assumption of linearity mapping between the bending angle of front segment of endoscope and motor rotation angle. 

\subsection{EndoVLA-Motion Dataset}
Given the absence of multimodal data for continuum robots, we construct EndoVLA-Motion, a vision-language and kinematic dataset, to train our EndoVLA model. EndoVLA-Motion dataset comprises 6k image–action pairs across three tasks: polyp tracking, abnormal region localization, and circular-marker following. All data are acquired with the robotic endoscope system using two types of stomach phantoms embedding polyps, abnormal mucosal regions, and concentric arrays of black markers. Raw video streams are captured under teleoperated control and automatically annotated with bounding boxes via YOLOv5~\cite{yolo}, followed by manual curation to remove erroneous detections. Each bounding box is mapped to one of four discrete bending motions—upper‐right, upper‐left, lower‐left, or lower‐right—yielding the corresponding action label. As shown in Fig. \ref{fig:EndoVLA-fig}, the circular shape of focus region (FR) is defined. The center of FR is image center. The threshold of the desired observation region is set to be 28 pixels. If the distance between the center of the target and the center is less than the threshold, the motion type is modified to be ``still". We split the EndoVLA dataset into training (80\%) and evaluation (20\%) partitions to ensure robust model assessment. The prompts consist of two crucial components: (1) a detailed visual scene description generated by the Qwen2-VL base model, and (2) an explicit formulation of the spatial relationship between the localized target and the corresponding optimal bending motion direction. This structured prompt design enables the model to establish clear correlations between visual inputs, spatial reasoning, and appropriate robotic motion commands.

\section{Vision-Language-Action Instruction Tuning}
\subsection{Preliminaries}

\noindent Our VLA model is built upon the Qwen2-VL backbone. We jointly process the current RGB observation $O_t \in \mathbb{R}^{H\times W\times 3}$ and the instruction prompt \(I\) to produce the next control command. At each timestep \(t\), the current RGB frame \(O_t\) is mapped into a latent vector via a frozen visual encoder \(E_{v}\), while the instruction \(I\) is tokenized and embedded by a fixed language encoder \(E_{l}\).  These two feature vectors are then combined and passed through the finetuned language model \(f_{\phi}\). The output tokens is given by
$T_l \;=\; f_{\phi}\!\bigl(E_{v}(O_t),\,E_{l}(I)\bigr),$ where $T_l$ is the decoded text.

\subsection{EndoVLA Model}

\textbf{Architecture.} Instead of full parameter training, we employ Low-Rank Adaptation (LoRA) \cite{lora} to efficiently fine-tune the LLM while keeping most of the base model parameters frozen.
For a given RGB image $O_t$ and a language instruction $I$, the model predicts both a bounding box $B_t$ of the target and an appropriate action $A_t$. This can be formulated as $(B_t, A_t) = f_{\phi}(O_t, I).$
The RGB image processing pipeline involves a fixed vision encoder that extracts visual features, followed by a trainable MLP projector that aligns these features with the language embedding space of the LLM. The language instruction is processed directly by the LLM's tokenizer. This multimodal fusion allows the model to understand both the visual context and the linguistic commands, enabling it to make accurate predictions for the tracking tasks.
The bounding box prediction $B_t$ is represented in the format $[x, y, w, h]$ where $(x, y)$ denotes the top-left corner coordinates and $(w, h)$ represents the width and height of the box. The action $A_t$ corresponds to one of the predefined discrete actions.

\textbf{Problem Formation.} Given an instruction \(I\) that specifies the target name, the LLM must first transform this linguistic input into an object detection task on the RGB image to localize the target, and then decide whether to actuate the robot or hold position based on the localization result. Let \(p_t = (u_t, v_t)^\top \in \mathbb{R}^2\) denote the pixel coordinates of the detected target center in the current image frame, and \(p_f = (u_f, v_f)^\top \in \mathbb{R}^2\) denote the pixel coordinates of the desired focus region center.  Denote by \(\boldsymbol{\theta} = [\theta_1,\theta_2]^\top\in\mathbb{R}^2\) the incremental rotations of the two actuation motors (one pair for bending towards right/left, one pair for bending towards up/down), and assume each bending angle \(\alpha_i\) is linearly related by \(\alpha_i = k\,\theta_i\) for some constant \(k\). 
We define the (nonlinear) mapping from motor rotations to image coordinates of the target as $p_t =\mathcal{M}(\boldsymbol{\theta})\,, \  
\mathcal{M}:\mathbb{R}^2\to\mathbb{R}^2.$
We define a discrete action set $\mathcal{A} = \{\text{upper-right},\,\text{upper-left},\,\text{lower-left},\,\text{lower-right},\,\text{stop}\},$
each non-stop action \(a\in \mathcal{A} \) corresponding to a fixed motor increment
\(\Delta\boldsymbol{\theta}(a)\):
\begin{equation}
\begin{aligned}
\Delta\boldsymbol{\theta}(\text{upper-right})   &= [+\,\delta\theta_1,\;+\,\delta\theta_2]^\top,\quad
\Delta\boldsymbol{\theta}(\text{lower-left}) &= [-\,\delta\theta_1,\;-\,\delta\theta_2]^\top, \\
\Delta\boldsymbol{\theta}(\text{upper-left}) &= [-\,\delta\theta_1,\;+\delta\theta_2]^\top,\quad
\Delta\boldsymbol{\theta}(\text{lower-right})&= [+\,\delta\theta_1,\;-\,\delta\theta_2]^\top,
\end{aligned}
\label{eq:theta_directions}
\end{equation}

and the special “stop” action is executed when
\(\|p_t - p_f\|_2 \le \epsilon\), indicating mission success. The $\epsilon$ is a fixed value, which is defined when the training dataset is built. We set $\epsilon = 18$  pixels. A vision–language action LLM parameterized by \(\phi\) maps the
current image and instruction \(I\) to a discrete action: $a_t =f_{\phi}\!\bigl(E_v(O_t),\,E_l(I)\bigr)
\in\mathcal{A}.$
When ``stop'' signal has not yet been issued, applying \(a_t\) for updating the motors: $ \boldsymbol{\theta}_{t+1}
=\boldsymbol{\theta}_t+\Delta\boldsymbol{\theta}(a_t),$
and the new target position is $p_{t+1}=\mathcal{M}\bigl(\boldsymbol{\theta}_{t+1}\bigr).$

The control objective is to choose actions so that the pixel‐distance
\(\|p_t - p_f\|_2\) is reduced below a threshold \(\epsilon\) as quickly
as possible, and then issue “stop”:
\begin{equation}
a_t^*
=\;\begin{cases}
\displaystyle
\arg\min_{a\in \mathcal{A}}
\Bigl\|\,
\mathcal{M}\bigl(\boldsymbol{\theta}_t + \Delta\boldsymbol{\theta}(a)\bigr)
- p_f
\Bigr\|_2,
& \|p_t - p_f\|_2 > \epsilon, \\[1ex]
\text{stop}, & \|p_t - p_f\|_2 \le \epsilon.
\end{cases}
\label{eq:optimal_action}
\end{equation}

\textbf{Input.} EndoVLA processes two input components: an RGB frame $O_{t}$ captured by the eye-in-hand endoscopic camera, and a prompt $I$ that combines a textual description $I_{T}$ with a motion guideline $I_{G}$. To generate $I_{T}$, we leverage the image captioning capabilities of the pretrained LLM to describe the visual scene. For $I_{G}$, we incorporate in-context learning principles by providing exemplars that demonstrate the reasoning process for selecting appropriate actions based on target localization \cite{ICL}. Specifically, we include examples that show how to reason about the spatial relationship between the detected target and the desired focus point, followed by the corresponding action selection.

\subsection{Dual-phase Fine-tuning (DFT)}
\textbf{Supervised Fine-tuning.} We implement a LoRA adaptor~\cite{lora} to align visual representations with the LLM enabling effective processing of endoscopic images with language instructions. The training objective minimizes prediction errors on ground-truth bounding boxes and action prediction.

\textbf{Reinforcement Learning with Verifiable Rewards.} To further enhance the model's performance beyond supervised learning, we employ Reinforcement Learning with Verifiable Rewards. This approach uses reward signals that can be objectively computed from the model's predictions rather than relying on subjective human preferences or black-box reward models.

The optimization is across different target–focus configurations (groups), and we employ Group Relative Policy Optimization (GRPO) \cite{grpo}. Let \(\mathcal{G}\) be the set of groups (i.e.\ different target positions or image contexts) and for each group \(g\in\mathcal{G}\) let trajectories \(\tau^g=(s_0,a_0,\dots)\) be collected under the old policy \(\pi_{\phi_{\mathrm{old}}}\).  Define the group‐specific importance ratio and advantage:
\begin{equation}
r_t^g(\phi) \;=\;\frac{\pi_{\phi}(a_t\mid s_t, g)}{\pi_{\phi_{\mathrm{old}}}(a_t\mid s_t, g)},
\quad
A_t^g \;=\;R_t^g - V_{\psi}(s_t, g),
\label{eq:ratio_advantage}
\end{equation}
where \(R_t^g\) is the cumulative reward (e.g.\ IoU, MA, Format) for group \(g\), and \(V_{\psi}(s_t,g)\) is a value function estimate.
The GRPO surrogate objective with clipping and a KL penalty across groups is:
\begin{align}
L^{\mathrm{GRPO}}(\phi)
&= \sum_{g\in\mathcal{G}} w_g \;\mathbb{E}_{t\sim\tau^g}\Bigl[
    \min\bigl(r_t^g(\phi)\,A_t^g,\,
    \mathrm{clip}(r_t^g(\phi),1-\epsilon,1+\epsilon)\,A_t^g\bigr)
\Bigr] \nonumber\\
&\quad - \,\alpha \sum_{g\in\mathcal{G}} w_g\;\mathbb{E}_{t\sim\tau^g}\Bigl[
    \mathrm{KL}\bigl[\pi_{\phi_{\mathrm{old}}}(\cdot\mid s_t,g)\,\big\|\,\pi_{\phi}(\cdot\mid s_t,g)\bigr]
\Bigr].
\end{align}

where \(w_g\) are nonnegative group‐weighting coefficients, \(\epsilon\) is the clipping threshold, and \(\alpha\) scales the KL penalty. The policy parameters are updated by
$\phi^* = \arg\max_{\phi}\;L^{\mathrm{GRPO}}(\phi)\,,$ ensuring that each group’s policy improves relative to its own baseline while maintaining stability across updates.

In our framework, verifiable rewards provide quantitative feedback for the model's predictions based on direct measurement of prediction quality. We designed three complementary reward functions:
1. \textbf{IoU Reward}: This reward measures the Intersection over Union between the predicted bounding box and the ground truth bounding box:
\begin{equation}
\label{eq:IoU}
R_{IoU}(B_{\text{pred}}, B_{\text{gt}}) = \frac{\text{Area}(B_{\text{pred}} \cap B_{\text{gt}})}{\text{Area}(B_{\text{pred}} \cup B_{\text{gt}})}
\end{equation}
where $B_{\text{pred}} = [x_{\text{pred}}, y_{\text{pred}}, w_{\text{pred}}, h_{\text{pred}}]$ is the predicted bounding box and $B_{\text{gt}} = [x_{\text{gt}}, y_{\text{gt}}, w_{\text{gt}}, h_{\text{gt}}]$ is the ground truth bounding box.

2. \textbf{Motion Angle (MA) Reward}: This binary reward evaluates whether the predicted action matches the ground truth action:
\begin{equation}
R_{MA}(A_{\text{pred}}, A_{\text{gt}}) = 
\begin{cases}
1.0 & \text{if } A_{\text{pred}} = A_{\text{gt}} \\
0.0 & \text{otherwise}
\end{cases}
\end{equation}
where actions are selected from a discrete action set $\mathcal{A}$ defined at Eq. \ref{eq:theta_directions}.

3. \textbf{Format Reward}: This reward ensures the model's output adheres to the required format of $[x, y, w, h]a_t$, where the data type of discrete action $a_t$ is character:
\begin{equation}
R_{Format}(output) = 
\begin{cases}
1.0 & \text{if output matches } [\text{digit}, \text{digit}, \text{digit}, \text{digit}] \; \text{character} \\
0.0 & \text{otherwise}
\end{cases}
\end{equation}

\section{Experiments}
\label{sec:exp}

\subsection{Implementation Details}

Our experiments address two key questions: (1) Can EndoVLA accurately track targets in dynamic endoluminal environments using RGB images and textual prompts with limited training data? (2) Does explicit object localization improve motion prediction compared to direct motion commands?

EndoVLA is first SFT using Unsloth~\cite{unsloth} and then RFT with VLM-R1~\cite{vlm}. Our base model is Qwen2-VL-7B~\cite{qwen2}, with LoRA-based DFT on an NVIDIA A6000 GPU. The training of our model on the EndoVLA-motion dataset was performed for a single epoch. For the SFT phase, we employed the AdamW optimizer with an initial learning rate of 2e-4 and implemented a linear decay schedule, completing the training in approximately 3 hours with a batch size of 2. Subsequently, during the RFT phase, we maintained the AdamW optimizer but reduced the initial learning rate to 1e-5, utilizing a batch size of 4 and a group size of 4 (g=4 in Eq. \ref{eq:ratio_advantage}) for 4 hours training.

\begin{figure}[t]
  \centering
  \includegraphics[width=\linewidth]{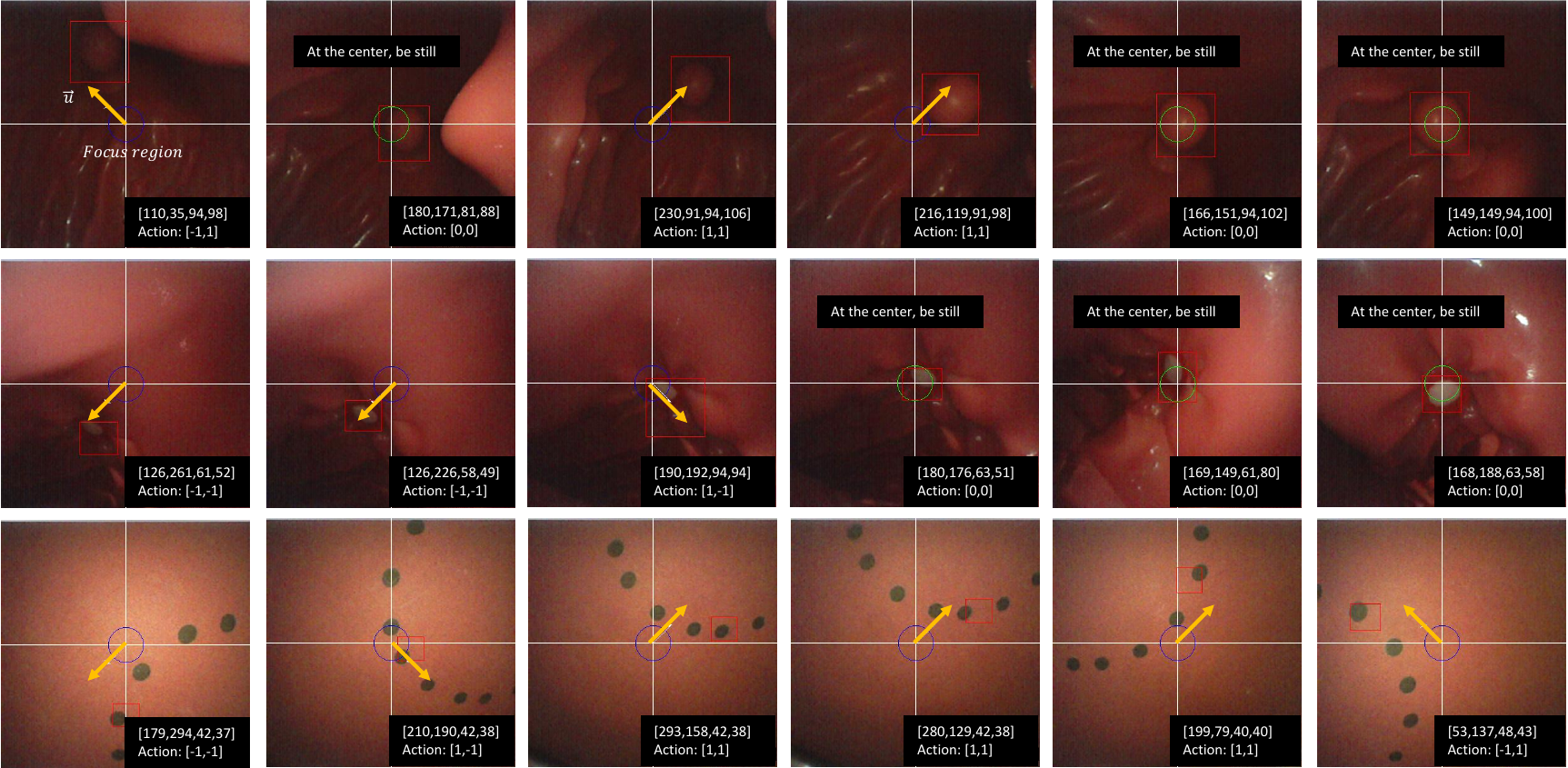}
  \caption{Example of successful rollouts on the real-world endoscopic tasks (PP, AR and CC) on a robotic endoscope setup and stomach phantoms.}
  \label{fig3}
\end{figure}

\begin{figure}[t]
  \centering
  \includegraphics[width=\linewidth]{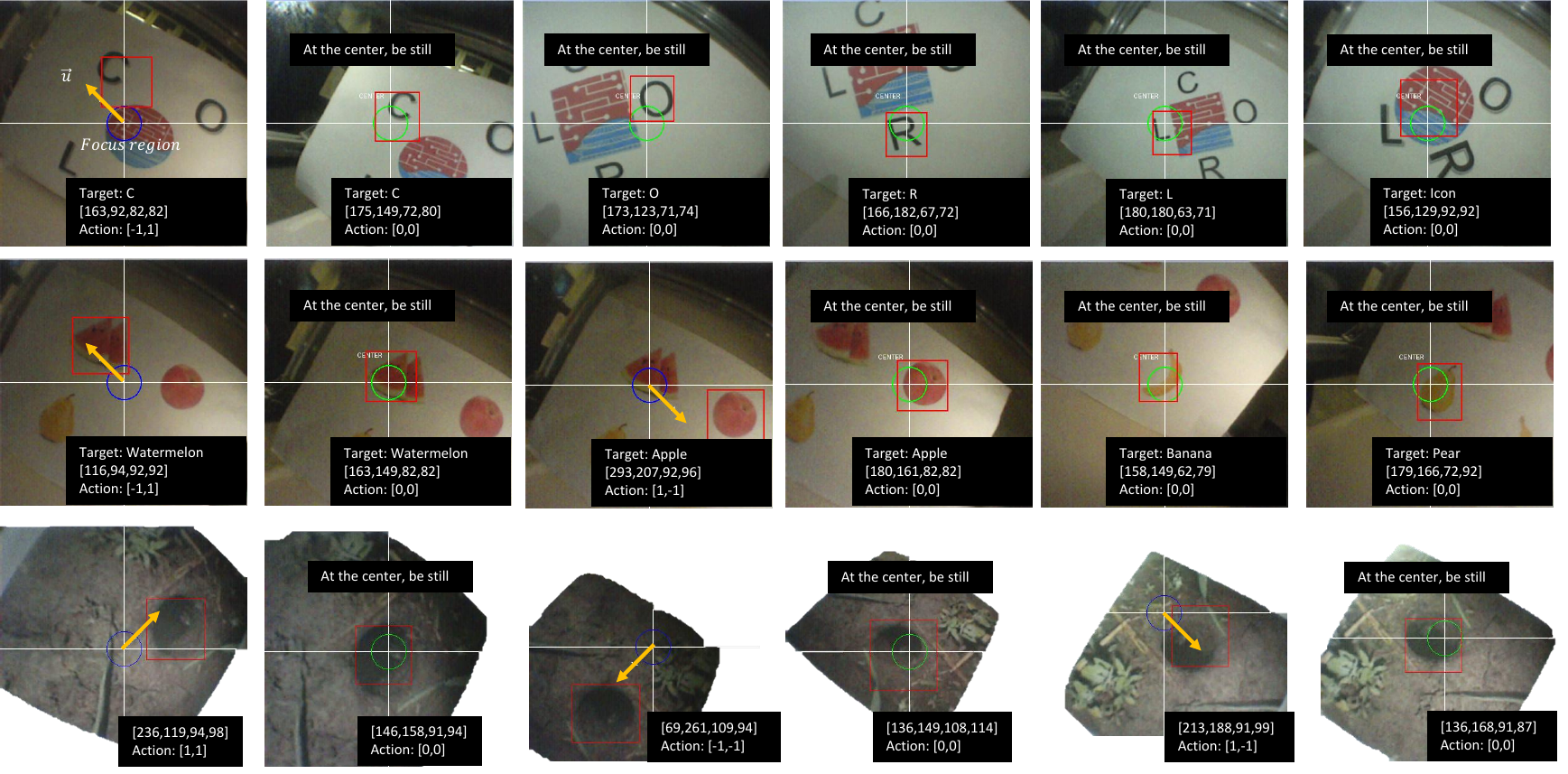}
  \caption{Example of successful rollouts on a real-world robotic setup of tasks: CORL character and icon sequential tracking, fruit sequence tracking, and hole tracking in the field.}
  \label{fig4.pdf}
\end{figure}

We evaluate three endoscopic tasks (Figure \ref{fig3}): polyp tracking (PP), abnormal region tracking (AR), and circular cutting marker following (CC), with increasing localization difficulty. We test three instruction types (Figure \ref{fig:EndoVLA-fig}): bounding box localization ($I_b$), directional localization ($I_d$), and direct action prediction ($I_a$). Additionally, we assess generalization in non-endoscopic scenes (Figure \ref{fig4.pdf}): hole tracking, sequential character tracking (CORL), and sequential fruit tracking.

\subsection{Motion Prediction with Base Model}
\begin{wraptable}{r}{0.55\textwidth}

\vspace{-15pt}
\setlength\tabcolsep{0.2em}
\caption{\textbf{Motion Type Prediction of Base Model}}
\vspace{-0.5em} 
\centering
\renewcommand{\arraystretch}{0.5} 
\begin{tabular}{@{}l|cc|cc|cc@{}}
\toprule
\textbf{Tasks} & 
{\makecell{PP/$I_a$}} & 
{\makecell{PP/$I_d$}} & 
{\makecell{AR/$I_a$}} & 
{\makecell{AR/$I_d$}} & 
{\makecell{CC/$I_a$}} & 
{\makecell{CC/$I_d$}} \\ \midrule
Random               &  20.0 &  20.0 &  20.0 &  20.0 &  25.0 &  25.0 \\
$\text{Img}_{raw}$   &  35.8 & 44.0  & 36.5  & 31.5  & \textbf{26.5} & 20.8 \\
$\text{Img}_{gt}$    &  \textbf{38.4} & \textbf{50.2} & \textbf{42.1} & \textbf{45.1} & 18.2 & \textbf{24.7} \\
$\uparrow$ (\%)      &  7.3  & 14.1  & 15.3  & 43.2  & -31.3 & 18.8 \\
\bottomrule
\end{tabular}
\vspace{-0.5em} 
\label{tab:evalbase}
\vspace{-10pt}
\end{wraptable}


We evaluated the base Qwen2-VL model using instructions $I_d$ and $I_a$ on both raw images ($\text{Img}_{raw}$) and images with ground truth bounding boxes ($\text{Img}_{gt}$). Table~\ref{tab:evalbase} shows that localization instructions ($I_d$) improve performance, particularly for PP. Adding ground-truth boxes under $I_d$ yielded improvements of +7.3\% (PP), +43.2\% (AR), and +18.8\% (CC), confirming that pairing explicit localization with reliable bounding boxes boosts motion prediction.

\setlength\tabcolsep{0.2em}
\begin{table}[ht]
\vspace{-1.5em} 
\caption{\textbf{IoU of bounding box and PR (\%) of action prediction}}
\centering
\renewcommand{\arraystretch}{0.4} 
\begin{tabular}{@{}c|ccc|ccc|ccc@{}}
\toprule
\textbf{Model}  
& \makecell{PP\\(IoU, $I_b$)}
& \makecell{PP\\($I_b$)}
& \makecell{PP\\($I_a$)}
& \makecell{AR\\(IoU, $I_b$)}
& \makecell{AR\\($I_b$)}
& \makecell{AR\\($I_a$)}
& \makecell{CC\\(IoU, $I_b$)}
& \makecell{CC\\($I_b$)}
& \makecell{CC\\($I_a$)} \\
\midrule
SFT           & \textbf{62.1}  & \textbf{87.7} & \textbf{66.4} & 45.7  & \textbf{81.5} & \textbf{70} & \textbf{11.0}  & 57.1  & 26.8   \\
RFT       & 58.0  & 73.5 & 57.6  & \textbf{48.2} & 61.3 & 34.9 & 2.8  & \textbf{65.5}  & \textbf{34.4}   \\
\midrule
SFT + RFT               & \textbf{86.1}  & \textbf{94.8} & \textbf{78.7} & \textbf{79.1}  & \textbf{92.0} & \textbf{85} & \textbf{48.5}  & \textbf{89.7}   & \textbf{56}   \\
$\uparrow$ (\%)      & 38.6 & 8.1 & 18.5 & 64.1 & 12.9 & 21.4 & 340.9 & 36.9  & 62.8 \\
\bottomrule
\end{tabular}
\vspace{-1em} 
\label{tab:ioump}
\end{table}

\subsection{Ablation Study}

Table~\ref{tab:ioump} presents IoU for bounding box predictions and precision rate (PR) for motion predictions across different FT strategies. Single-phase approaches showed task-specific strengths: SFT performed well on PP (87.7\% with $I_b$) but struggled with complex localization in CC (11.0\% IoU), while RFT showed complementary strengths in CC but underperformed in PP.

Our DFT approach demonstrated superior performance across all metrics and tasks, with IoU improvements of 38.6\% (PP), 64.1\% (AR), and 340.9\% (CC) compared to single-phase approaches. Motion prediction accuracy reached 94.8\% (PP), 92.0\% (AR), and 89.7\% (CC) under $I_b$, representing improvements of 8.1\%, 12.9\%, and 36.9\% respectively.

\begin{table*}[ht]
\caption{\textbf{Performance in real-world endoscopic tracking tasks.}}
\centering
\begin{minipage}{0.62\textwidth}
\setlength\tabcolsep{0.2em}
\centering
\renewcommand{\arraystretch}{0.5} 
\begin{tabular}{@{}l|cccc|cccc@{}}
\toprule
\textbf{Model} & 
\makecell{PP\\($I_b$, c)} & 
\makecell{PP\\($I_b$, r)} & 
\makecell{PP\\($I_a$, c)} & 
\makecell{PP\\($I_a$, r)} & 
\makecell{AR\\($I_b$, c)} & 
\makecell{AR\\($I_b$, r)} & 
\makecell{AR\\($I_a$, c)} & 
\makecell{AR\\($I_a$, r)} 
\\ \midrule
SFT   & 70.0 & 33.3 & 50.0 & 30.0 & 83.0 & 40.0 & 57.0 & 20.0 \\ 
RFT   & 37.0 & 37.0 & 43.0 & 0.0  & 0.0  & 0.0  & 0.0  & 0.0  \\ 
\midrule
SFT+RFT   & \textbf{100.0} & \textbf{100.0} & \textbf{100.0} & \textbf{63.0} & \textbf{100.0} & \textbf{100.0} & \textbf{100.0} & \textbf{57.0} \\
\bottomrule
\end{tabular}
\subcaption{\textbf{SR (\%) of PP and AR tracking tasks}}
\vspace{-1em} 
\label{tab:SR_tracking}
\end{minipage}
\hfill
\begin{minipage}{0.34\textwidth}
\setlength\tabcolsep{0.2em}
\centering
\renewcommand{\arraystretch}{0.5} 
\begin{tabular}{@{}lcc@{}}
\toprule
\textbf{Model} & 
\makecell{CR\\($I_b$)} &
\makecell{SR\\($I_b$)} \\
\midrule
SFT   & 50.0 & 0.0 \\ 
RFT   & 45.0 & 0.0 \\ 
\midrule
SFT+RFT   & \textbf{66.7} & \textbf{10.0} \\
\bottomrule
\end{tabular}
\subcaption{\textbf{CR (\%) and SR (\%) of CC task}}
\vspace{-1em} 
\label{tab:SR_CC}
\end{minipage}
\vspace{-1em} 
\end{table*}

\subsection{Real Robot Evaluation}

For PP and AR, the maximum steps are 30 across 30 trials to position the targets within the FR. We measured SR using two metrics: moving closer (\textbf{c}) and moving within FR (\textbf{r}). For CC, models had 200 steps across 10 trials, with completion rate (CR) and SR measurements.
The DFT approach achieved 100\% success in moving toward targets in both PP and AR tasks, with significant improvements in precise positioning (63.0\% for PP and 57.0\% for AR with $I_a$), outperforming single-phase approaches by 110.0\% and 185.0\%, respectively. For the challenging CC task, only the DFT model achieved any success (10.0\%) in completing the entire circle.
\setlength
\tabcolsep{0.17em}
\begin{table}[ht]
\caption{\textbf{SR (\%) in general scene tracking tasks}} 
\renewcommand{\arraystretch}{0.5} 
\centering
\begin{tabular}{@{}lc|ccccc|c|cccc|c@{}}
\toprule
\multirow{2}{*}{\textbf{Model}} & \multicolumn{6}{c|}{\textbf{CORL Character Sequence}} & \multicolumn{5}{c}{\textbf{Fruit Sequence}} & \multirow{2}{*}{\textbf{Hole}} \\
\cmidrule(lr){2-7} \cmidrule(lr){8-12}
& \textbf{Full} & \textbf{$C_{FR}$} & \textbf{$O_{FR}$} & \textbf{$R_{FR}$} & \textbf{$L_{FR}$} & \textbf{$Icon_{FR}$} & \textbf{Full} & 
\textbf{$W_{FR}$} & \textbf{$A_{FR}$} & \textbf{$P_{FR}$} & \textbf{$B_{FR}$} & \\
\midrule
SFT      & 0.0 & 0.0 & 0.0 & 10.0 & 0.0 & 0.0 & 90.0 & 100.0& 100.0 & 100.0 & 90.0 & 30.0 \\

RFT      & 0.0 & 0.0 & 0.0 & 0.0 & 0.0 & 0.0 & 0.0 & 0.0 & 0.0 & 0.0 & 0.0 & 0.0 \\
\midrule
SFT+RFT  & \textbf{50.0} & \textbf{100.0} & \textbf{70.0} & \textbf{60.0} & \textbf{60.0} & \textbf{80.0} & \textbf{100.0} & \textbf{100.0} & \textbf{100.0} & \textbf{100.0} & \textbf{100.0} & \textbf{90.0} \\
\bottomrule
\end{tabular}
\label{tab:general}
\end{table}

\subsection{Generalization to General Scenes}

We evaluated generalization to non-endoscopic tasks: CORL character and icon tracking, fruit sequence tracking, and hole tracking in the field. The maximum number of steps across 10 trials is 200, as measured by the SR. As shown in Table~\ref{tab:general}, the DFT approach demonstrated remarkable zero-shot performance, achieving a 50\% success rate for the complete CORL sequence, 100\% for fruit tracking, and 90\% success in tracking hole in the field. In contrast, the RFT-only model generated excessively large bounding boxes and showed inconsistent motion predictions, while the SFT model struggled with visually different scenes (30\% success in tracking hole).

These results confirm that the DFT enhances both endoscopic performance and zero-shot generalization by developing robust localization capabilities that transfer effectively across visual contexts.
\section{Conclusion}

In this work, we present EndoVLA, a novel Vision-Language-Action model tailored for autonomous tracking in endoscopic procedures using continuum robots. Addressing the critical challenges of limited endoscopic data and domain-specific generalization, we introduce a DFT strategy that combines SFT and RFT guided by verifiable, task-specific rewards. We construct the EndoVLA-Motion dataset, a vision-language-kinematics dataset, to facilitate training under realistic robotic settings. Our extensive experiments across both endoscopic and non-endoscopic tasks demonstrate that EndoVLA not only achieves state-of-the-art performance in precise target localization and motion prediction but also exhibits robust zero-shot generalization to novel environments and task sequences. This work establishes a foundation for more autonomous, precise robotic assistance in VLA for continuum robots.
\bibliography{ref}  

\begin{thebibliography}{40}
\providecommand{\natexlab}[1]{#1}
\providecommand{\url}[1]{\texttt{#1}}
\expandafter\ifx\csname urlstyle\endcsname\relax
  \providecommand{\doi}[1]{doi: #1}\else
  \providecommand{\doi}{doi: \begingroup \urlstyle{rm}\Url}\fi

\bibitem[Gao et~al.(2024)Gao, Yang, Xiao, Zhu, Zhang, Hou, Liu, Meng, Sun, Zuo, et~al.]{gaoijrr}
H.~Gao, X.~Yang, X.~Xiao, X.~Zhu, T.~Zhang, C.~Hou, H.~Liu, M.~Q.-H. Meng, L.~Sun, X.~Zuo, et~al.
\newblock Transendoscopic flexible parallel continuum robotic mechanism for bimanual endoscopic submucosal dissection.
\newblock \emph{The International Journal of Robotics Research}, 43\penalty0 (3):\penalty0 281--304, 2024.

\bibitem[Maple et~al.(2015)Maple, Dayyeh, Chauhan, Hwang, Komanduri, Manfredi, Konda, Murad, Siddiqui, and Banerjee]{maple2015endoscopic}
J.~T. Maple, B.~K.~A. Dayyeh, S.~S. Chauhan, J.~H. Hwang, S.~Komanduri, M.~Manfredi, V.~Konda, F.~M. Murad, U.~D. Siddiqui, and S.~Banerjee.
\newblock Endoscopic submucosal dissection.
\newblock \emph{Gastrointestinal endoscopy}, 81\penalty0 (6):\penalty0 1311--1325, 2015.

\bibitem[Lee et~al.(2021)Lee, Cheon, Kim, and Kwon]{lee2021easyendo}
D.-H. Lee, B.~Cheon, J.~Kim, and D.-S. Kwon.
\newblock easyendo robotic endoscopy system: Development and usability test in a randomized controlled trial with novices and physicians.
\newblock \emph{The International Journal of Medical Robotics and Computer Assisted Surgery}, 17\penalty0 (1):\penalty0 1--14, 2021.

\bibitem[Zhang et~al.(2020)Zhang, Ly, Nithyanand, Modayil, Khodorskiy, Neppala, Bhumi, DeMaria, Widmer, Friedel, et~al.]{zhang2020learning}
X.~Zhang, E.~K. Ly, S.~Nithyanand, R.~J. Modayil, D.~O. Khodorskiy, S.~Neppala, S.~Bhumi, M.~DeMaria, J.~L. Widmer, D.~M. Friedel, et~al.
\newblock Learning curve for endoscopic submucosal dissection with an untutored, prevalence-based approach in the united states.
\newblock \emph{Clinical Gastroenterology and Hepatology}, 18\penalty0 (3):\penalty0 580--588, 2020.

\bibitem[Lin et~al.(2023)Lin, Li, Chu, Dou, Liu, and Au]{lin2023end}
H.~Lin, B.~Li, X.~Chu, Q.~Dou, Y.~Liu, and K.~W.~S. Au.
\newblock End-to-end learning of deep visuomotor policy for needle picking.
\newblock In \emph{2023 IEEE/RSJ International Conference on Intelligent Robots and Systems (IROS)}, pages 8487--8494. IEEE, 2023.

\bibitem[Hwang et~al.(2020)Hwang, Seita, Thananjeyan, Ichnowski, Paradis, Fer, Low, and Goldberg]{hwang2020applying}
M.~Hwang, D.~Seita, B.~Thananjeyan, J.~Ichnowski, S.~Paradis, D.~Fer, T.~Low, and K.~Goldberg.
\newblock Applying depth-sensing to automated surgical manipulation with a da vinci robot.
\newblock In \emph{2020 international symposium on medical robotics (ISMR)}, pages 22--29. IEEE, 2020.

\bibitem[Islam et~al.(2019)Islam, Li, and Ren]{islam2019learning}
M.~Islam, Y.~Li, and H.~Ren.
\newblock Learning where to look while tracking instruments in robot-assisted surgery.
\newblock In \emph{International Conference on Medical Image Computing and Computer-Assisted Intervention}, pages 412--420. Springer, 2019.

\bibitem[Leonard et~al.(2018)Leonard, Sinha, Reiter, Ishii, Gallia, Taylor, and Hager]{leonard2018evaluation}
S.~Leonard, A.~Sinha, A.~Reiter, M.~Ishii, G.~L. Gallia, R.~H. Taylor, and G.~D. Hager.
\newblock Evaluation and stability analysis of video-based navigation system for functional endoscopic sinus surgery on in vivo clinical data.
\newblock \emph{IEEE transactions on medical imaging}, 37\penalty0 (10):\penalty0 2185--2195, 2018.

\bibitem[Fu et~al.(2021)Fu, Jin, Zhang, He, Zha, Hu, Gan, Yan, Wang, and Ye]{fu2021future}
Z.~Fu, Z.~Jin, C.~Zhang, Z.~He, Z.~Zha, C.~Hu, T.~Gan, Q.~Yan, P.~Wang, and X.~Ye.
\newblock The future of endoscopic navigation: a review of advanced endoscopic vision technology.
\newblock \emph{IEEE Access}, 9:\penalty0 41144--41167, 2021.

\bibitem[Long et~al.(2023)Long, Wei, Huang, Wang, and Dou]{long2023human}
Y.~Long, W.~Wei, T.~Huang, Y.~Wang, and Q.~Dou.
\newblock Human-in-the-loop embodied intelligence with interactive simulation environment for surgical robot learning.
\newblock \emph{IEEE Robotics and Automation Letters}, 8\penalty0 (8):\penalty0 4441--4448, 2023.

\bibitem[Xu et~al.(2021)Xu, Li, Lu, Liu, Dou, and Heng]{xu2021surrol}
J.~Xu, B.~Li, B.~Lu, Y.-H. Liu, Q.~Dou, and P.-A. Heng.
\newblock Surrol: An open-source reinforcement learning centered and dvrk compatible platform for surgical robot learning.
\newblock In \emph{2021 IEEE/RSJ International Conference on Intelligent Robots and Systems (IROS)}, pages 1821--1828. IEEE, 2021.

\bibitem[Huang et~al.(2023)Huang, Chen, Li, Liu, and Dou]{huang2023guided}
T.~Huang, K.~Chen, B.~Li, Y.-H. Liu, and Q.~Dou.
\newblock Guided reinforcement learning with efficient exploration for task automation of surgical robot.
\newblock In \emph{2023 IEEE International Conference on Robotics and Automation (ICRA)}, pages 4640--4647. IEEE, 2023.

\bibitem[Li et~al.(2024)Li, Wang, Dai, Ma, Ng, Hu, and Li]{li2024robonurse}
S.~Li, J.~Wang, R.~Dai, W.~Ma, W.~Y. Ng, Y.~Hu, and Z.~Li.
\newblock Robonurse-vla: Robotic scrub nurse system based on vision-language-action model.
\newblock \emph{arXiv preprint arXiv:2409.19590}, 2024.

\bibitem[Moghani et~al.(2024)Moghani, Doorenbos, Panitch, Huver, Azizian, Goldberg, and Garg]{moghani2024sufia}
M.~Moghani, L.~Doorenbos, W.~C.-H. Panitch, S.~Huver, M.~Azizian, K.~Goldberg, and A.~Garg.
\newblock Sufia: language-guided augmented dexterity for robotic surgical assistants.
\newblock In \emph{2024 IEEE/RSJ International Conference on Intelligent Robots and Systems (IROS)}, pages 6969--6976. IEEE, 2024.

\bibitem[Achiam et~al.(2023)Achiam, Adler, Agarwal, Ahmad, Akkaya, Aleman, Almeida, Altenschmidt, Altman, Anadkat, et~al.]{achiam2023gpt}
J.~Achiam, S.~Adler, S.~Agarwal, L.~Ahmad, I.~Akkaya, F.~L. Aleman, D.~Almeida, J.~Altenschmidt, S.~Altman, S.~Anadkat, et~al.
\newblock Gpt-4 technical report.
\newblock \emph{arXiv preprint arXiv:2303.08774}, 2023.

\bibitem[Liu et~al.(2023)Liu, Li, Wu, and Lee]{liu2023visual}
H.~Liu, C.~Li, Q.~Wu, and Y.~J. Lee.
\newblock Visual instruction tuning.
\newblock \emph{Advances in neural information processing systems}, 36:\penalty0 34892--34916, 2023.

\bibitem[Wang et~al.(2025)Wang, Bai, Wang, Yuan, Li, Jiang, He, Wu, Chen, Lei, et~al.]{wang2025endochat}
G.~Wang, L.~Bai, J.~Wang, K.~Yuan, Z.~Li, T.~Jiang, X.~He, J.~Wu, Z.~Chen, Z.~Lei, et~al.
\newblock Endochat: Grounded multimodal large language model for endoscopic surgery.
\newblock \emph{arXiv preprint arXiv:2501.11347}, 2025.

\bibitem[Zitkovich et~al.(2023)Zitkovich, Yu, Xu, Xu, Xiao, Xia, Wu, Wohlhart, Welker, Wahid, et~al.]{zitkovich2023rt}
B.~Zitkovich, T.~Yu, S.~Xu, P.~Xu, T.~Xiao, F.~Xia, J.~Wu, P.~Wohlhart, S.~Welker, A.~Wahid, et~al.
\newblock Rt-2: Vision-language-action models transfer web knowledge to robotic control.
\newblock In \emph{Conference on Robot Learning}, pages 2165--2183. PMLR, 2023.

\bibitem[Liu et~al.(2024)Liu, Liu, Wang, An, Li, Zhou, Yang, Zhang, Guo, and Zhang]{liu2024robomamba}
J.~Liu, M.~Liu, Z.~Wang, P.~An, X.~Li, K.~Zhou, S.~Yang, R.~Zhang, Y.~Guo, and S.~Zhang.
\newblock Robomamba: Efficient vision-language-action model for robotic reasoning and manipulation.
\newblock \emph{Advances in Neural Information Processing Systems}, 37:\penalty0 40085--40110, 2024.

\bibitem[Arai et~al.(2025)Arai, Miwa, Sasaki, Watanabe, Yamaguchi, Aoki, and Yamamoto]{arai2025covla}
H.~Arai, K.~Miwa, K.~Sasaki, K.~Watanabe, Y.~Yamaguchi, S.~Aoki, and I.~Yamamoto.
\newblock Covla: Comprehensive vision-language-action dataset for autonomous driving.
\newblock In \emph{2025 IEEE/CVF Winter Conference on Applications of Computer Vision (WACV)}, pages 1933--1943. IEEE, 2025.

\bibitem[Kim et~al.(2024)Kim, Pertsch, Karamcheti, Xiao, Balakrishna, Nair, Rafailov, Foster, Lam, Sanketi, et~al.]{kim2024openvla}
M.~J. Kim, K.~Pertsch, S.~Karamcheti, T.~Xiao, A.~Balakrishna, S.~Nair, R.~Rafailov, E.~Foster, G.~Lam, P.~Sanketi, et~al.
\newblock Openvla: An open-source vision-language-action model.
\newblock \emph{arXiv preprint arXiv:2406.09246}, 2024.

\bibitem[Rudiman(2021)]{rudiman2021minimally}
R.~Rudiman.
\newblock Minimally invasive gastrointestinal surgery: from past to the future.
\newblock \emph{Annals of Medicine and Surgery}, 71:\penalty0 102922, 2021.

\bibitem[Fan et~al.(2024)Fan, Chen, Ferrigno, and De~Momi]{fan2024learn}
K.~Fan, Z.~Chen, G.~Ferrigno, and E.~De~Momi.
\newblock Learn from safe experience: Safe reinforcement learning for task automation of surgical robot.
\newblock \emph{IEEE Transactions on Artificial Intelligence}, 5\penalty0 (7):\penalty0 3374--3383, 2024.

\bibitem[Brohan et~al.(2022)Brohan, Brown, Carbajal, Chebotar, Dabis, Finn, Gopalakrishnan, Hausman, Herzog, Hsu, et~al.]{brohan2022rt}
A.~Brohan, N.~Brown, J.~Carbajal, Y.~Chebotar, J.~Dabis, C.~Finn, K.~Gopalakrishnan, K.~Hausman, A.~Herzog, J.~Hsu, et~al.
\newblock Rt-1: Robotics transformer for real-world control at scale.
\newblock \emph{arXiv preprint arXiv:2212.06817}, 2022.

\bibitem[O’Neill et~al.(2024)O’Neill, Rehman, Maddukuri, Gupta, Padalkar, Lee, Pooley, Gupta, Mandlekar, Jain, et~al.]{o2024open}
A.~O’Neill, A.~Rehman, A.~Maddukuri, A.~Gupta, A.~Padalkar, A.~Lee, A.~Pooley, A.~Gupta, A.~Mandlekar, A.~Jain, et~al.
\newblock Open x-embodiment: Robotic learning datasets and rt-x models: Open x-embodiment collaboration 0.
\newblock In \emph{2024 IEEE International Conference on Robotics and Automation (ICRA)}, pages 6892--6903. IEEE, 2024.

\bibitem[Hu et~al.(2023)Hu, Xie, Jain, Francis, Patrikar, Keetha, Kim, Xie, Zhang, Fang, et~al.]{hu2023toward}
Y.~Hu, Q.~Xie, V.~Jain, J.~Francis, J.~Patrikar, N.~Keetha, S.~Kim, Y.~Xie, T.~Zhang, H.-S. Fang, et~al.
\newblock Toward general-purpose robots via foundation models: A survey and meta-analysis.
\newblock \emph{arXiv preprint arXiv:2312.08782}, 2023.

\bibitem[Schmidgall et~al.(2024)Schmidgall, Kim, Kuntz, Ghazi, and Krieger]{schmidgall2024general}
S.~Schmidgall, J.~W. Kim, A.~Kuntz, A.~E. Ghazi, and A.~Krieger.
\newblock General-purpose foundation models for increased autonomy in robot-assisted surgery.
\newblock \emph{Nature Machine Intelligence}, pages 1--9, 2024.

\bibitem[Shridhar et~al.(2023)Shridhar, Manuelli, and Fox]{shridhar2023perceiver}
M.~Shridhar, L.~Manuelli, and D.~Fox.
\newblock Perceiver-actor: A multi-task transformer for robotic manipulation.
\newblock In \emph{Conference on Robot Learning}, pages 785--799. PMLR, 2023.

\bibitem[Brohan et~al.(2023)Brohan, Brown, Carbajal, Chebotar, Chen, Choromanski, Ding, Driess, Dubey, Finn, et~al.]{brohan2023rt}
A.~Brohan, N.~Brown, J.~Carbajal, Y.~Chebotar, X.~Chen, K.~Choromanski, T.~Ding, D.~Driess, A.~Dubey, C.~Finn, et~al.
\newblock Rt-2: Vision-language-action models transfer web knowledge to robotic control.
\newblock \emph{arXiv preprint arXiv:2307.15818}, 2023.

\bibitem[Black et~al.(2024)Black, Brown, Driess, Esmail, Equi, Finn, Fusai, Groom, Hausman, Ichter, et~al.]{black2024pi_0}
K.~Black, N.~Brown, D.~Driess, A.~Esmail, M.~Equi, C.~Finn, N.~Fusai, L.~Groom, K.~Hausman, B.~Ichter, et~al.
\newblock $\displaystyle \pi_0 $: A vision-language-action flow model for general robot control.
\newblock \emph{arXiv preprint arXiv:2410.24164}, 2024.

\bibitem[Qian and Ren(2025)]{qian2025deep}
C.~Qian and H.~Ren.
\newblock Deep reinforcement learning in surgical robotics: enhancing the automation level.
\newblock \emph{Handbook of Robotic Surgery}, pages 89--102, 2025.

\bibitem[Liu et~al.(2024)Liu, Andres, Jiang, Luo, Shu, and Tsaftaris]{liu2024surgical}
J.~Liu, A.~Andres, Y.~Jiang, X.~Luo, W.~Shu, and S.~A. Tsaftaris.
\newblock Surgical task automation using actor-critic frameworks and self-supervised imitation learning.
\newblock \emph{arXiv preprint arXiv:2409.02724}, 2024.

\bibitem[Kim et~al.(2024)Kim, Zhao, Schmidgall, Deguet, Kobilarov, Finn, and Krieger]{kim2024surgical}
J.~W. Kim, T.~Z. Zhao, S.~Schmidgall, A.~Deguet, M.~Kobilarov, C.~Finn, and A.~Krieger.
\newblock Surgical robot transformer (srt): Imitation learning for surgical tasks.
\newblock \emph{arXiv preprint arXiv:2407.12998}, 2024.

\bibitem[Redmon et~al.(2016)Redmon, Divvala, Girshick, and Farhadi]{yolo}
J.~Redmon, S.~Divvala, R.~Girshick, and A.~Farhadi.
\newblock You only look once: Unified, real-time object detection.
\newblock In \emph{Proceedings of the IEEE conference on computer vision and pattern recognition}, pages 779--788, 2016.

\bibitem[Hu et~al.(2022)Hu, Shen, Wallis, Allen-Zhu, Li, Wang, Wang, Chen, et~al.]{lora}
E.~J. Hu, Y.~Shen, P.~Wallis, Z.~Allen-Zhu, Y.~Li, S.~Wang, L.~Wang, W.~Chen, et~al.
\newblock Lora: Low-rank adaptation of large language models.
\newblock \emph{ICLR}, 1\penalty0 (2):\penalty0 3, 2022.

\bibitem[Dong et~al.(2022)Dong, Li, Dai, Zheng, Ma, Li, Xia, Xu, Wu, Liu, et~al.]{ICL}
Q.~Dong, L.~Li, D.~Dai, C.~Zheng, J.~Ma, R.~Li, H.~Xia, J.~Xu, Z.~Wu, T.~Liu, et~al.
\newblock A survey on in-context learning.
\newblock \emph{arXiv preprint arXiv:2301.00234}, 2022.

\bibitem[Shao et~al.(2024)Shao, Wang, Zhu, Xu, Song, Bi, Zhang, Zhang, Li, Wu, et~al.]{grpo}
Z.~Shao, P.~Wang, Q.~Zhu, R.~Xu, J.~Song, X.~Bi, H.~Zhang, M.~Zhang, Y.~Li, Y.~Wu, et~al.
\newblock Deepseekmath: Pushing the limits of mathematical reasoning in open language models.
\newblock \emph{arXiv preprint arXiv:2402.03300}, 2024.

\bibitem[Daniel~Han and team(2023)]{unsloth}
M.~H. Daniel~Han and U.~team.
\newblock Unsloth, 2023.
\newblock URL \url{http://github.com/unslothai/unsloth}.

\bibitem[Shen et~al.(2025)Shen, Liu, Li, Fang, Ma, Liao, Shen, Zhang, Zhao, Zhang, Xu, and Zhao]{vlm}
H.~Shen, P.~Liu, J.~Li, C.~Fang, Y.~Ma, J.~Liao, Q.~Shen, Z.~Zhang, K.~Zhao, Q.~Zhang, R.~Xu, and T.~Zhao.
\newblock Vlm-r1: A stable and generalizable r1-style large vision-language model.
\newblock \emph{arXiv preprint arXiv:2504.07615}, 2025.

\bibitem[Wang et~al.(2024)Wang, Bai, Tan, Wang, Fan, Bai, Chen, Liu, Wang, Ge, et~al.]{qwen2}
P.~Wang, S.~Bai, S.~Tan, S.~Wang, Z.~Fan, J.~Bai, K.~Chen, X.~Liu, J.~Wang, W.~Ge, et~al.
\newblock Qwen2-vl: Enhancing vision-language model's perception of the world at any resolution.
\newblock \emph{arXiv preprint arXiv:2409.12191}, 2024.

\end{thebibliography}
\section{Limitations}

\textbf{Dataset Size and Diversity.} The EndoVLA-Motion dataset is relatively small (about 6k image-action pairs) and limited to only three phantom-based tasks. This restricted dataset may constrain the model’s generalizability to a wider range of endoscopic scenarios. For example, we have not tested the model on real procedures with varied pathologies (such as active bleeding or mucosal irregularities) or on additional tasks that occur in clinical settings. The controlled, simplified nature of our training data (clean phantoms) may not capture the full complexity of in vivo conditions.

\textbf{Environmental and Visual Variability.} Our current evaluations use phantom models under ideal conditions. In contrast, real endoscopic environments often involve challenges like significant soft-tissue deformation, specular reflections, bleeding, smoke, mucus, or partial occlusions from surgical instruments. These factors can severely degrade visual tracking performance, yet they were not present in our experiments. As a result, the robustness of our model in such adverse visual conditions remains unverified.

\textbf{Motion Control and Actuation Assumptions.} The system uses a discrete set of motion commands (four bending directions and a stop) as a simplification of the continuum motion space. It also assumes a simple linear mapping between the predicted command and the actual motor rotation angle. These choices ignore complex factors such as nonlinear kinematics, actuation delays, or varying compliance in the endoscope mechanism. This coarse discretization and linear approximation may limit fine-grained manipulation precision and accuracy. 

\textbf{Inference Speed and Temporal Responsiveness.} The full EndoVLA model runs at roughly 2 Hz, which is significantly lower than video frame rates (30 Hz). This latency means that fast-moving targets could outpace the controller, potentially causing delayed reactions or overshooting. Moreover, our model currently processes each frame independently without explicit temporal context. It cannot anticipate future motion or utilize history, which further hinders performance in rapidly changing scenes.

\textbf{Observed Failure Modes.} We observed several task-specific failure cases in our experiments. 
\begin{itemize}
    \item In the circular cutting task, for instance, the EndoVLA sometimes tracked the target ring in the wrong rotational direction (clockwise instead of counterclockwise).
    \item In zero-shot general scene tests, the model occasionally misidentified targets (for example, confusing similar letters or shapes) and lost track of the intended object. These errors highlight challenges in disambiguating visual cues and following instructions when multiple plausible targets are present.
    \item In sequential tracking scenarios, our current approach lacks a mechanism for detecting or recovering when the next target falls outside the field of view after completing tracking of the current target. The system continues to issue motion commands without correction when tracking is lost or when model predictions become uncertain. We have not implemented confidence estimation or a re-localization strategy that would allow the system to recognize tracking failures and recover from them. This absence of uncertainty-aware control and recovery mechanisms means the system may fail silently in scenarios where tracking continuity is compromised.
\end{itemize}  

\textbf{Future Works.} While EndoVLA has demonstrated strong performance under controlled conditions, its path toward clinical deployment requires several key advancements. First, we will integrate temporal and uncertainty-aware modeling to enhance safety and robustness. By extending the model to incorporate sequence-based architectures and real‐time confidence estimation, the system can anticipate rapid target motion, detect out‐of‐distribution inputs, and trigger failsafe behaviors—such as pausing actuation or reverting to manual control—when uncertainty exceeds safe thresholds. We will also explore continuous control outputs, moving beyond the current discrete action set to predict continuous bending angles or velocities, and investigate model compression and hardware acceleration (e.g., distillation) to boost inference speed toward real‐time rates ($\geq$30 Hz).

Second, we plan to broaden the training data and evaluation scenarios to bridge the gap between phantom studies and the complexity of in vivo endoscopy. This includes collecting and annotating diverse clinical videos that capture bleeding, smoke, soft‐tissue deformation, and occlusions, as well as designing multi‐step and multi‐target tracking benchmarks. Complementing real data with simulated variations and domain‐adaptation techniques will help address data scarcity. Finally, we will conduct human‐in‐the‐loop user studies in wet‐lab environments to assess endoscopists' trust cognitive load, and procedural efficiency, comparing EndoVLA’s assistance against traditional methods, which will be essential to refine the system for safe and effective use in real endoscopic workflows.  
\clearpage
\begin{center}
    \LARGE \textbf{Supplementary Materials}\\[1.5ex]
\end{center}
\vspace{2ex}
\appendix
\section{Dataset Creation Pipeline}
\label{sec:appendix-dataset}

To construct the EndoVLA-Motion dataset, we developed an automated labeling pipeline using YOLOv5-based object detection, followed by human-in-the-loop curation. The dataset includes three task-specific tracking categories: (1) \textbf{PP}: polyp tracking, (2) \textbf{AR}: delineation and following of abnormal mucosal regions, and (3) \textbf{CC}: adherence to predefined circular markers during circumferential cutting.

\subsection{Data Collection Setup}
\begin{figure}[t]
  \centering
  \includegraphics[width=\linewidth]{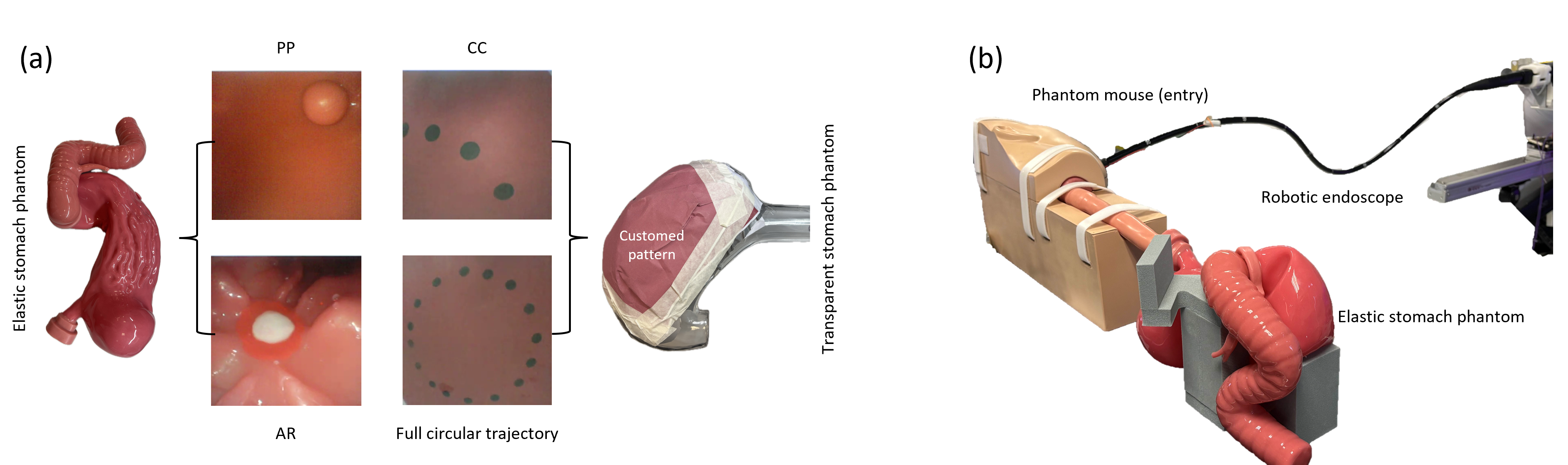}
  \caption{(a) The data is collected by two phantoms. (b) Robotic setup.}
  \label{fig:datacollection_setup}
\end{figure}

Figure~\ref{fig:datacollection_setup}a illustrates the physical setup used for data acquisition. Two different phantoms were employed to simulate gastrointestinal environments. Figure~\ref{fig:datacollection_setup}b shows the robotic setup, which includes a tendon-driven continuum endoscope manipulated by a motorized actuation unit, facilitating controlled and repeatable motions during data collection.

\subsection{Automated Labeling for PP and AR Tasks}
For both PP and AR tasks, frames were sampled from endoscopic videos (30 FPS). YOLOv5s, pre-trained on COCO, was employed to detect the target (e.g., polyp or abnormal region) in each selected frame. The highest-confidence detection bounding box $B = [x, y, w, h]$ was retained.

We defined a circular \textit{Focus Region} (FR) centered in the image with a radius threshold $r=20\sqrt{2}$ pixels (resolution is 400 $\times$ 400). The center of the bounding box was computed and compared to the FR center to determine motion labels:
\begin{itemize}
    \item If $\lVert \text{center}(B) - \text{center(FR)} \rVert < r$, the action label is [0,0] (no movement).
    \item Otherwise, the image is partitioned into four quadrants (upper-right, upper-left, lower-left, lower-right), and the label of motor angle ($\mathcal{A}_{MA}$) increment vector is assigned accordingly: $(1,1),(-1,1),(-1,-1),(1,-1)$.
\end{itemize}

Each selected frame is saved in three variants: raw, with central cross lines, and with annotated bounding boxes. The results are represented as \texttt{[x, y, w, h]$\mathcal{A}_{MA}$}.

\subsection{Automated Labeling for CC Task}
The task CC involves tracking a sequence of markers arranged in a circular pattern. To determine the next movement step, we first detect all visible bounding boxes using YOLOv5 with low thresholds ($\texttt{conf}=0.02$, $\texttt{iou}=0.04$), extract their centroids, and fit a circle via least-squares optimization.

We then select the next target point in anti-clockwise order relative to the current center-most marker. The corresponding bounding box is retrieved and the label of motor angle ($\mathcal{A}_{MA}$) increment vector is assigned accordingly: $(1,1),(-1,1),(-1,-1),(1,-1)$.

\subsection{Manual Curation and Finalization}

\begin{figure}[t]
  \centering
  \includegraphics[width=\linewidth]{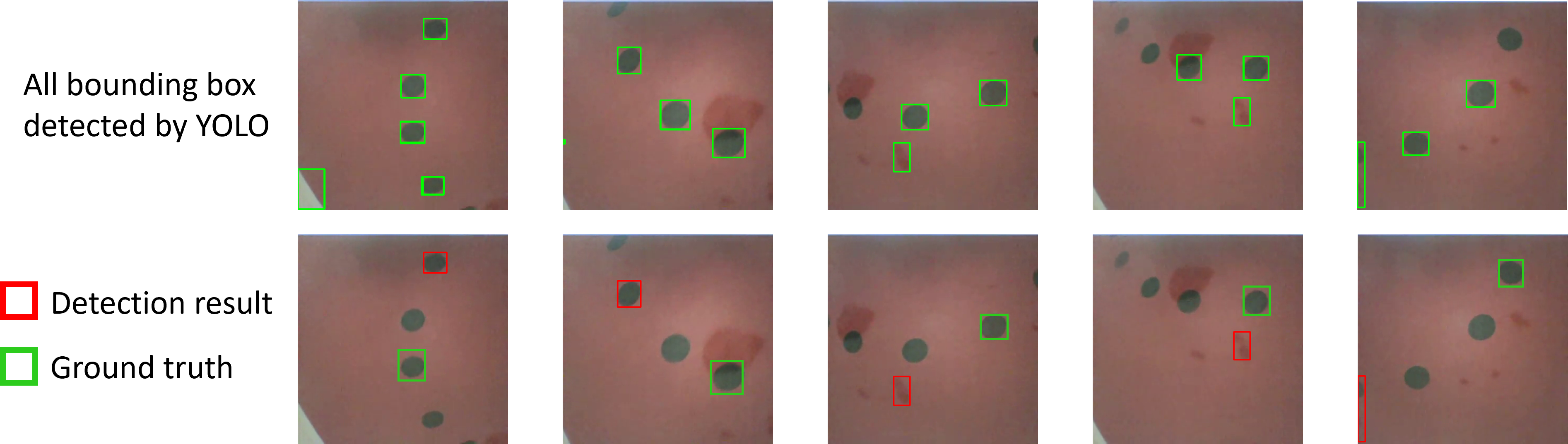}
  \caption{The examples for wrongly annotated the next tracking point in anti-clockwise direction.}
  \label{fig:appendixannotationpipeline}
\end{figure}

Figure~\ref{fig:appendixannotationpipeline} highlights examples of labeling failures encountered during the automated annotation of the CC task. Specifically, it shows cases where the next target point along the anti-clockwise circular trajectory was incorrectly. The labeling accuracy was approximately 65\% due to:
\begin{itemize}
    \item High sensitivity to detection noise,
    \item Overfitting to hyperparameters such as detection thresholds,
    \item Errors in angle sorting during circular fitting.
\end{itemize}

To ensure data quality, we manually reviewed all annotated frames and removed erroneous samples, yielding a high-confidence vision-language-kinematics dataset. The final curated dataset comprises 6k samples.

\section{EndoVLA-Motion Dataset}
\subsection{Prompt Design for Directional Action Labeling}
\label{sec:appendix-prompt}

We designed structured instruction prompts to enable the Vision-Language model to infer directional motion commands based on visual inputs. The prompts explicitly describe the target type and provide a consistent spatial-to-action mapping, allowing the model to learn task-aware control behaviors.

\paragraph{Instruction Format.} Each prompt is composed of:
\begin{itemize}
    \item A visual grounding prefix (\texttt{<image>}) to indicate the presence of image input;
    \item A task-specific target item's name (e.g., “sphere polyp” or “abnormal region”). The target is depicted by requesting qwen2-VL with prompt--What you see in the image?;
    \item A spatial-to-motion mapping that converts relative target location into one of five discrete MA action vectors:
    \begin{itemize}
        \item Upper right $\rightarrow$ [1, 1]
        \item Upper left $\rightarrow$ [-1, 1]
        \item Lower left $\rightarrow$ [-1, -1]
        \item Lower right $\rightarrow$ [1, -1]
        \item Center $\rightarrow$ [0, 0]
    \end{itemize}
\end{itemize}

\paragraph{Instruction Examples.}
We used three variants of prompts for each task:

\textbf{(1) Direct Action Output:}
\begin{itemize}
    \item \textbf{$I_a(PP)$:} “\texttt{<image> I want you to act as a direction leader. Please find the target that is a sphere polyp in the image. If the target is at lower left, output [-1,-1]; upper right, [1,1]; lower right, [1,-1]; upper left, [-1,1]; at center, [0,0]. Output action only.}”

\end{itemize}

\textbf{(2) Bounding box localization format:}
\begin{itemize}
    \item \textbf{$I_b(PP)$:} “\texttt{<image> I want you to act as a expert for object detection and direction leader. Please find the target that is a sphere polyp in the image. To determine the bounding box of target, define its left lower start point as (x,y), the width is w, the height is h. If the target is at lower left, output [-1,-1]; upper right, [1,1]; lower right, [1,-1]; upper left, [-1,1]; at center, [0,0]. Output bounding box and action.}”
\end{itemize}

\textbf{(3) Directional format:}
\begin{itemize}
    \item \textbf{$I_d(PP)$:} “\texttt{<image> I want you to act as a expert for direction leader. Please find the target that is a sphere polyp in the image. If the target is at lower left, output "lower left[-1,-1]"; upper right, "upper right[1,1]"; lower right, "lower right[1,-1]"; upper left, "upper left[-1,1]"; at center, "center[0,0]". Output direction and action.}”
\end{itemize}
\subsection{Dataset Annotation and Statistics}
\setlength\tabcolsep{0.2em}
\begin{table}[H]
\caption{\textbf{Type and number of the motion annotation}} 
\centering
\begin{tabular}{@{}lcccccccc}
\toprule
\textbf{MA}        & \textbf{[1,1]} & \textbf{[-1,1]} & \textbf{[-1,-1]} & \textbf{[1,-1]} & \textbf{[0,0]} & \textbf{All}\\ \midrule
TS (PP)                     & 373           & 69             & 270           & 216   &  125          &  1053       \\
TS (AR)                     & 67            & 236            & 493           & 202   &  132          &  1130       \\
TS (CC)                     & 747           & 458            & 804           & 824   &  N/A          &  2833       \\
ES (PP)                     & 113           & 21             & 61            & 48    &  36           &  279        \\
ES (AR)                     & 16            & 62             & 107           & 40    &  41           &  266        \\
ES (CC)                     & 191           & 131            & 201           & 195   &  N/A          &  718        \\
All Number                  & 1507          & 977            & 1936          & 1525  &   334         &  6279       \\
\bottomrule
\end{tabular}
\label{tab:motiontype}
\end{table}

The total number of annotation sample $N_{t}$ is 6279. Robotic endoscope enable four bending motion types: upper right, upper left, lower left, and lower right. The dataset is split into 80\% training set (TS) and 20\% evaluation set (ES) and still when object is within the FR. Table \ref{tab:motiontype} summarizes the distribution. We trained the model with two prompt type: $I_a$ and $I_b$. The total number of samples is 2 $\times$ $N_{t}$ = 12558.

\label{sec:appendix-grounding}
\begin{figure}[t]
  \centering
  \includegraphics[width=\linewidth]{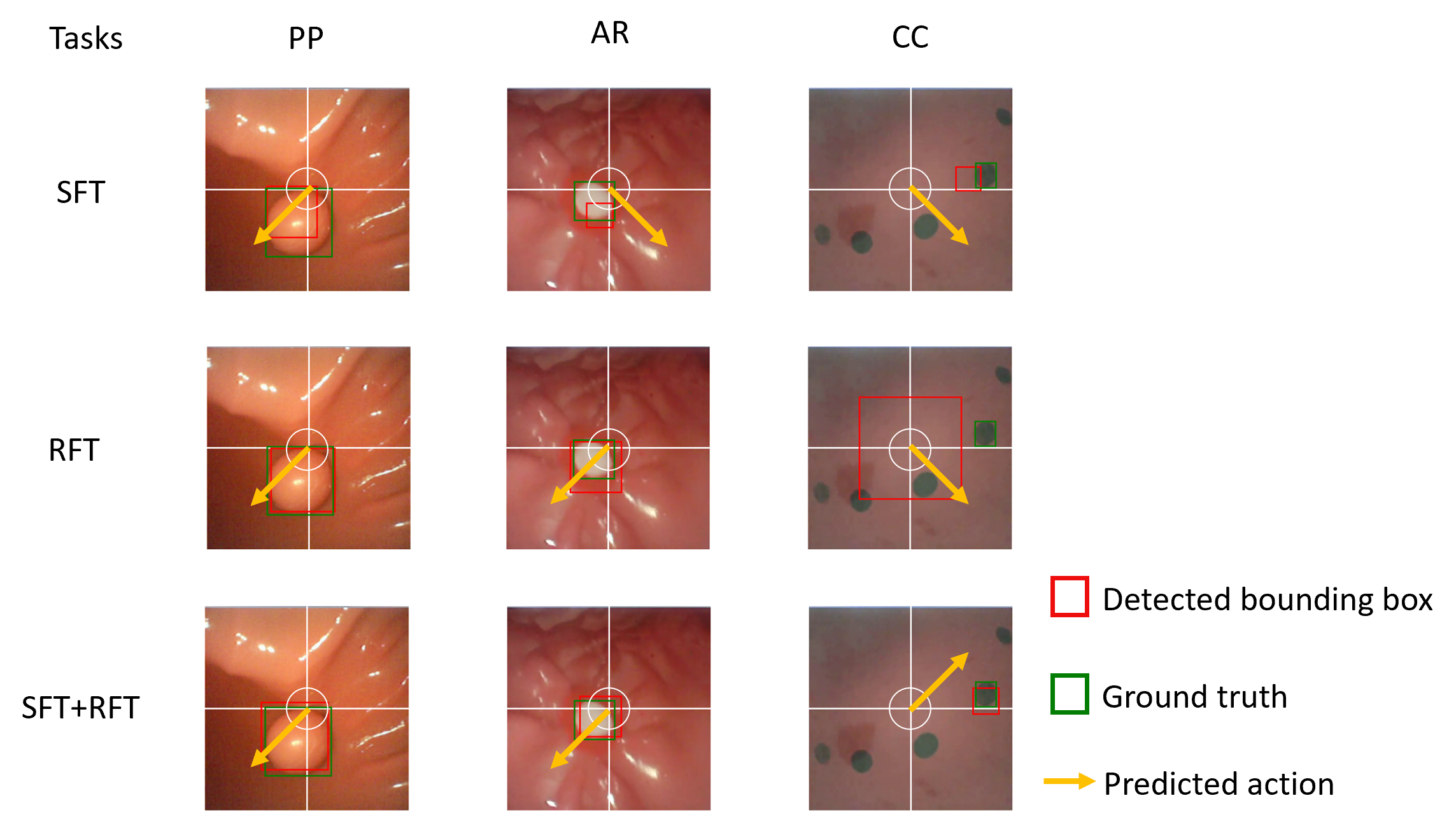}
  \caption{Qualitative results of bounding box prediction (green: GT, red: prediction) and motion output (yellow arrows) across the three tasks: PP, AR and CC. Top to bottom (fine-tuning method): SFT, RFT, SFT+RFT.}
  \vspace{-2em} 
  \label{fig:grounding-qual}
\end{figure}
To evaluate the grounding performance of different training paradigms, we present comparisons in Figure~\ref{fig:grounding-qual}. Each sub-image displays the predicted bounding box and action for a given task and training setting. 

\section{Grounding Capability across SFT, RFT, and DFT}

\textbf{Observation.}
\begin{itemize}
    \item \textbf{SFT} generally predicts well-formed bounding boxes, especially in simpler visual scenes (e.g., polyp tracking). However, action prediction can be suboptimal.
    \item \textbf{RFT} fails to localize objects accurately as the bounding box is too large.
    \item \textbf{SFT+RFT} synergizes both strengths, yielding correct target localization and precise directional action across all tasks (PP, AR, and CC).
\end{itemize}

\end{document}